\documentclass[sigconf]{acmart}
\setcopyright{none}
\renewcommand\footnotetextcopyrightpermission[1]{}
\acmDOI{}
\acmISBN{}
\usepackage{listings}
\usepackage{multirow}
\usepackage{graphics}
\usepackage{graphicx}
\usepackage{adjustbox}

\AtBeginDocument{%
  \providecommand\BibTeX{{%
    \normalfont B\kern-0.5em{\scshape i\kern-0.25em b}\kern-0.8em\TeX}}}

\acmConference[LegalIR (ECIR ’23)]{Make sure to enter the correct
  conference title from your rights confirmation emai}{April 02, 2023}{Dublin, Ireland}
\begin{document}

\title{Exploring Semi-supervised Hierarchical Stacked Encoder for Legal Judgement Prediction}


\author{Nishchal Prasad}
\affiliation{%
  \city{IRIT, Toulouse}
  \country{France}}
\email{Nishchal.Prasad@irit.fr}

\author{Mohand Boughanem}
\affiliation{%
  \city{IRIT, Toulouse}
  \country{France}
}
\email{Mohand.Boughanem@irit.fr}

\author{Taoufiq Dkaki}
\affiliation{%
 \city{IRIT, Toulouse}
 \country{France}}
 \email{Taoufiq.Dkaki@irit.fr}


\begin{abstract}
  Predicting the judgment of a legal case from its unannotated case facts is a challenging task. The lengthy and non-uniform document structure poses an even greater challenge in extracting information for decision prediction. In this work, we explore and propose a two-level classification mechanism; both supervised and unsupervised; by using domain-specific pre-trained BERT to extract information from long documents in terms of sentence embeddings further processing with transformer encoder layer and use unsupervised clustering to extract hidden labels from these embeddings to better predict a judgment of a legal case. We conduct several experiments with this mechanism and see higher performance gains than the previously proposed methods on the ILDC dataset. Our experimental results also show the importance of domain-specific pre-training of Transformer Encoders in legal information processing. 
\end{abstract}






\maketitle
\section{Introduction}
Automating legal case proceedings can assist the decision-making process with speed and robustness, which can save time and be beneficial to both the legal authorities and the people concerned. One of the underlying tasks which deal with this broader aspect is the prediction of the outcome for the legal cases with just the facts of the case, which depicts the general real-life setting where only the case facts are provided. For this problem, many techniques have been explored in the past using machine learning to predict the outcome of legal cases.

For their Case Judgment Prediction and Explanation (CJPE) task, Malik et al. \cite{ildc-cjpe} introduced the Indian Legal Document Corpus (ILDC) dataset which reflects our ideal general setting for legal case documents. We use this dataset for testing our methods and compare them with other state-of-the-art models on the same. 
In our past work \cite{CIRCLE2022}, we 
demonstrated that a domain-specific pre-trained model can perform noticeably better and adapt effectively to domains of the same kind with different syntax, lexicon, and grammatical settings. Shounak et al.\cite{InLegalBERT} pre-trained BERT on a large corpus of Indian legal documents and applied it to several benchmark legal NLP tasks over both Indian legal text and those belonging to other countries. 
One problem with a BERT-based transformer architecture 
is the constraint in processing large documents due to the input limit of 512 tokens. In this work, we aim to predict decisions from large and non-uniform structured legal documents having very low annotations (i.e. just the prediction label). We explore the effects of some of the available legal (i.e domain-specific) pre-trained BERT models with an unsupervised clustering algorithm
(HDBSCAN\cite{hdbscan}) and propose a method, that leverages both of these techniques to understand long and unstructured legal case documents. 

\section{Method}
\begin{figure}[h] 
  \centering
  \includegraphics[width=\linewidth]{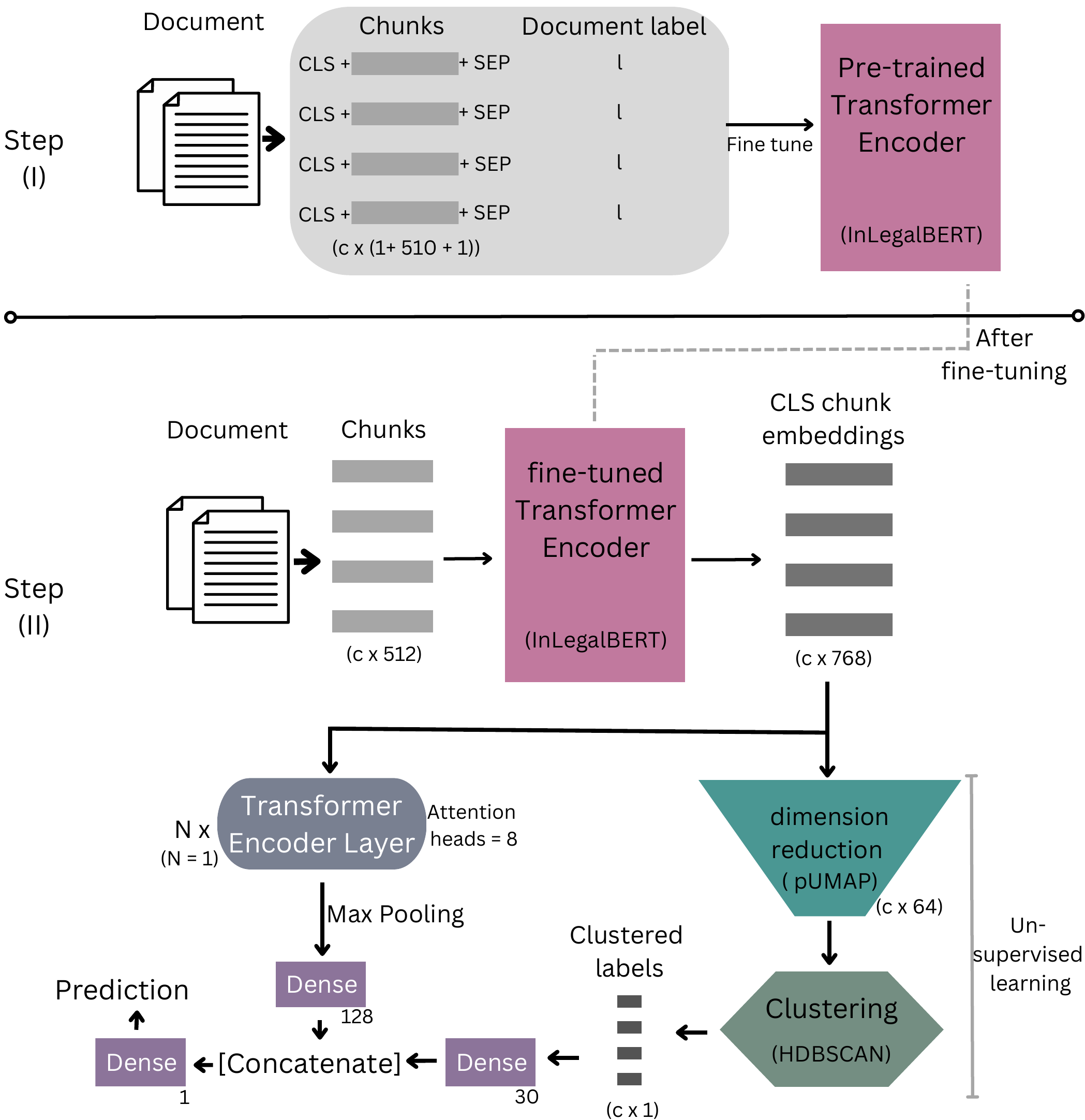}
  \caption{Two-level classification architecture}
  \label{fig:proposed_model}
\end{figure}

We modify the method of Hierarchical Transformer\cite{Hierarchical_Transformers} to tackle this problem of large document processing with the use of clustering to be able to extract more information for further processing. We experiment with two domain-specific pre-trained BERT models (LEGAL-BERT\cite{legal-bert} and InLegalBERT\cite{InLegalBERT}) with the hypothesis that domain pre-training of a transformer model is necessary 
for the in-domain vocabulary and lexical understanding \cite{CIRCLE2022}. We process the documents in two steps (figure \ref{fig:proposed_model}). 
We divide the document into parts called chunks (sequential sets of words). We tokenize and wrap these chunks with the [CLS] and [SEP] tokens. These tokenized chunks with their respective document label individually form input to a BERT model for fine-tuning (step I,  fig.  \ref{fig:proposed_model}).
 After fine-tuning, the [CLS] token embeddings are extracted for individual chunks which are used for the next step of processing.
The [CLS] embeddings are considered here to be a representation of the chunk, and concatenating them together gives an approximate representation of the whole document. 

 In step II, We use transformer layers on the extracted [CLS] embeddings for the inter-chunk attention to learn the whole document representation. We also experiment with RNNs (BiLSTM, GRU) after the transformer encoder (Table \ref{results}). 
The [CLS] embeddings are also used for the unsupervised learning mechanism i.e. clustering the individual chunks which are used as extra information while training. This provides information for the unlabeled parts of the document, i.e. which chunk relates to which topic. These individual cluster features along with the chunk embeddings help a model to better understand its contents and also add the constituent information of the related and unrelated parts of the document. For example, two chunks relating to the same law article in two different documents will be grouped together while clustering, and this grouping will be used as a piece of extra information extracted from the document. We have experimented with two variants of the inputs to the Transformer Encoder layer: The [CLS] chunk embeddings extracted from the finetuned BERT ($\alpha$), or the dimension-reduced [CLS] chunk embeddings ($\beta$) from pUMAP\footnote{\url{https://umap-learn.readthedocs.io/en/latest/parametric_umap.html}} having 64 dimensions. Table \ref{results} shows the impact of these two combinations on the classification performance. For clustering, we use HDBSCAN\cite{hdbscan} with a minimum cluster and sample size of 15 and 10 respectively.




\begin{table}
\centering
\caption{Experimental results of legal text classification on ILDC dataset for different architectures}
\begin{adjustbox}{width=\columnwidth}
\label{results}
\begin{tabular}{|c|c|c|c|c|c|c|c|} \hline
\multicolumn{2}{|c|}{\multirow{3}{*}{\begin{tabular}[c]{@{}c@{}}\textbf{Models} \\(e = epochs)\end{tabular}}} & \multicolumn{6}{c|}{Metrics (\%)~} \\ \cline{3-8}
\multicolumn{2}{|c|}{} & \multicolumn{3}{c|}{Validation} & \multicolumn{3}{c|}{Test} \\ \cline{3-8}
\multicolumn{2}{|c|}{} & Acc. & mP & mR & Acc. & mP & mR \\ \hline
\multicolumn{8}{|l|}{\textbf{Pre-Trained Transformer Encoders} (fine-tuned)} \\ \hline
\multicolumn{2}{|c|}{BERT\cite{CIRCLE2022}~~~e=2} & - & - & - & 60.52 & 66.13 & 60.55 \\ \hline
\multicolumn{2}{|c|}{XLNet\cite{CIRCLE2022}~~~e=2} & - & - & - & 70.51 & 72.01 & 70.09 \\ \hline
\multicolumn{2}{|c|}{LEGAL-BERT\cite{CIRCLE2022}~~~e=2} & - & - & - & 73.83 & 73.90 & 73.84 \\ \hline
\multicolumn{2}{|c|}{InLEGAL-BERT~ ~e=4} & 76.15 & 76.87 & 76.16 & 76.00 & 76.17 & 76.02 \\ \hline
\multicolumn{2}{|c|}{InLEGAL-BERT + BiGRU \cite{InLegalBERT}} & - & - & - & - & 83.43 & 83.15 \\ \hline
\multicolumn{8}{|l|}{\textbf{Two-level Architectures:~}} \\ \hline
\multirow{2}{*}{\begin{tabular}[c]{@{}c@{}}LEGAL-BERT+\\($fine$-$tuned$)\\$e=4$\end{tabular}} & \begin{tabular}[c]{@{}c@{}}Bi-LSTM + Dropout\\~e=6~~\cite{CIRCLE2022}\end{tabular} & - & - & - & 80.60 & 0.8106 & 80.63 \\ \cline{2-8}
 & \begin{tabular}[c]{@{}c@{}}Bi-LSTM + Dropout \\+Multi-head attention$_\beta$ \\~~~~e=6~~\cite{CIRCLE2022}\end{tabular} & - & - & - & 80.90 & 81.60 & 80.90 \\ \hline
\multirow{3}{*}{\begin{tabular}[c]{@{}c@{}}InLEGAL-BERT+ \\($fine$-$tuned$) \\$e=4$\end{tabular}} & $2\times$Bi-GRU e=3 & 83.37 & 83.35 & 83.28 & 83.31 & 83.39 & 83.30 \\ \cline{2-8}
 & Bi-LSTM + Bi-GRU e=3 & 83.97 & 83.40 & 83.25 & 83.11 & 83.76 & 83.09 \\ \cline{2-8}
 & $1\times$~Encoder e=3 & 84.10 & 84.33 & 84.10 & \textbf{83.72} & \textbf{83.74} & \textbf{83.72} \\ \hline
\multirow{3}{*}{\begin{tabular}[c]{@{}c@{}}InLEGAL-BERT+\\($fine$-$tuned$)\\$e=4$\\+pUMAP+HDBSCAN\\\end{tabular}} & $(\alpha)$~~~~~~$1\times$Encoder e=1 & \textbf{84.51} & \textbf{84.56} & \textbf{84.51} & \textbf{83.65} & \textbf{83.66} & \textbf{83.65} \\ \cline{2-8}
 & $(\beta)$~~~~~~$1\times$ Encoder e=1 & 83.90 & 84.01 & 83.90 & 83.39 & 83.39 & 83.39 \\ \cline{2-8}
 & \begin{tabular}[c]{@{}c@{}}$(\alpha)$~~~~~~$1\times$ Encoder \\+ BiLSTM e=3\end{tabular} & \multicolumn{1}{l|}{\textbf{85.01}} & \multicolumn{1}{l|}{\textbf{85.03}} & \textbf{85.01} & 83.59 & 83.59 & 83.58 \\ \hline
\end{tabular}
\end{adjustbox}
\end{table}

\section{Dataset and Results}
To conduct the experiments, we used the Indian Legal Document Corpus ILDC\cite{ildc-cjpe} to replicate a real-life setting of decision prediction of legal documents, for our proposed method. The dataset consists of $39898$ case proceedings (in English) from the Supreme Court of India (SCI). Each document is identified with the initial judgment rendered by the SCI judge(s) between 'rejected' and 'accepted'. Hence, our task of decision prediction can be formulated as a binary text classification problem. The dataset is pre-divided into a test (1517 documents) and validation (994 documents) set, we use the same for our experiments. 

We show concise results in Table \ref{results} amongst the experiments conducted with different architectures.
We used accuracy, macro-precision (mP), and macro-recall (mR), as the performance metrics and compare them with the previous baseline models. The InLegalBERT with RNNs performs 3 points higher than LEGAL-BERT, showing the effectiveness of further in-domain pre-training.  
The RNNs give almost the same performance as the Transformer Encoder layers in the test set, but the Encoders were more stable while training by showing marginal variations ($\approx 0.1$) in the validation metrics for a set of 3-4 subsequent epochs. Thus we chose the encoder layers to further learn from the [CLS] chunk embeddings. 
The effect of the unsupervised clustering mechanism with its combinations with the Transformer Encoder Layers, both inclusive and exclusive, can be seen in Table \ref{results}. The clustered information gives the model more features to learn from and increases performance in the validation set. Though the performance in the test set is not affected as much. This is because the clustering algorithm here is only trained on the train and validation set and not on the test set which affects the clusters on new data points (test set). Adding BiLSTM over the transformer encoder slightly affected the performance with an increase in the metrics for validation and a slight decrease in the test set. Most of the performance gain comes from the transformer encoder layer which helps the chunk [CLS] embeddings to attend to each other giving the overall document representation, while the cluster labels provide a few extra hidden features to improve the performance slightly. The footnote\footnote{\url{https://github.com/NishchalPrasad/Semi-supervised-Stacked-Encoder.git}} contains the code used for these experiments.

\section{conclusion}
This work introduces a framework to classify large unstructured legal documents using both a supervised and unsupervised learning mechanism achieving higher metrics on the experiments on the ILDC dataset over the previous baseline architectures. We demonstrate the effect of including features generated from an unsupervised clustering mechanism and see some relative gain with the same. We aim to explore further to extract the explanation of these predictions in the future and also develop methods to learn from long sequences.

\begin{acks}
This work is supported by LAWBOT project (ANR-20-CE38-0013) and HPC/AI resources from GENCI-IDRIS (2022-AD011013937).
\end{acks}

\bibliographystyle{ACM-Reference-Format}
\bibliography{bib_ecir23_abstract}

\end{document}